\documentclass[conference]{IEEEtran}
\IEEEoverridecommandlockouts
\usepackage{cite}
\usepackage{amsmath,amssymb,amsfonts}
\usepackage{algorithmic}
\usepackage{graphicx}
\usepackage{textcomp}
\usepackage{xcolor}
\def\BibTeX{{\rm B\kern-.05em{\sc i\kern-.025em b}\kern-.08em
    T\kern-.1667em\lower.7ex\hbox{E}\kern-.125emX}}
\begin{document}

\title{Robotic Imitation of Human Actions\\
{\footnotesize \textsuperscript{}}
\thanks{The authors gratefully acknowledge support from the DFG (CML, MoReSpace,
LeCAREbot), BMWK (SIDIMO, VERIKAS), and the European Commission
(TRAIL, TERAIS). }
}

\author{\IEEEauthorblockN{1\textsuperscript{st} Josua Spisak}
\IEEEauthorblockA{\textit{Knowledge Technology Group} \\
\textit{University of Hamburg}\\
Hamburg, Germany \\
josua.spisak@uni-hamburg.de}
\and
\IEEEauthorblockN{2\textsuperscript{nd} Matthias Kerzel}
\IEEEauthorblockA{\textit{Hamburger Informatik} \\
\textit{Technologie-Center}\\
Hamburg, Germany \\
matthias.kerzel@uni-hamburg.de}
\and
\IEEEauthorblockN{3\textsuperscript{rd} Stefan Wermter}
\IEEEauthorblockA{\textit{Knowledge Technology Group} \\
\textit{University of Hamburg}\\
Hamburg, Germany \\
stefan.wermter@uni-hamburg.de}
}
\maketitle

\begin{abstract}
Imitation can allow us to quickly gain an understanding of a new task. Through a demonstration, we can gain direct knowledge about which actions need to be performed and which goals they have. In this paper, we introduce a new approach to imitation learning that tackles the challenges of a robot imitating a human, such as the change in perspective and body schema. Our approach can use a single human demonstration to abstract information about the demonstrated task, and use that information to generalise and replicate it. We facilitate this ability by a new integration of two state-of-the-art methods: a diffusion action segmentation model to abstract temporal information from the demonstration and an open vocabulary object detector for spatial information. Furthermore, we refine the abstracted information and use symbolic reasoning to create an action plan utilising inverse kinematics, to allow the robot to imitate the demonstrated action.
\end{abstract}


\section{Introduction}
Imitation learning allows us to directly learn from the demonstrations of an expert to quickly adapt to new tasks and carry out complex chains of action. How exactly imitation learning works, or even what imitation is, can at times be hard to define. To gain a better understanding of how imitation learning works, Tomasello et al.\cite{tomasello1996apes} observe naturally occurring instances. Seemingly important for imitation learning in apes is to learn the specific sequence of actions to reach a desired goal, while the exact way in which, for example, the fingers are moving is less important \cite{whiten2004apes, byrne1998learning}.
Following the exact motions of an expert can be almost impossible just because of how each individual differs. When it comes to humanoid robots, this can be further exaggerated. Our robot {NICOL} (Neuro-Inspired COLlaborator) \cite{KASFHEW24}, for example, has more joints in its arm than a human. Therefore, following an example from a human demonstration by directly copying the joint movements would not make sense. However, following the same actions that led to the desired result is still very feasible. Closely connected to imitation learning is the famous mirror neuron concept \cite{gallese1998mirror}, where Gallese et al. found out that seeing certain actions can have very similar neuron activations as doing these actions. They specifically mention four actions where this pattern can be found: grasping an object, releasing an object, manipulating an object and tearing an object. These fundamental actions can be combined into more complex higher-level actions. Playing basketball is a sequence of grasping the ball, moving it and releasing it. So, if we are able to use a given body scheme to perform these actions and combine it with imitation learning to learn the sequence and goals of the action, it can allow us to imitate high-level actions \cite{byrne1998learning}. For these reasons, imitation learning has been a field of study in computer science for a long time \cite{schaal1999imitation}.\\
\begin{figure}[t]
    \centering
    \includegraphics[width=0.5\textwidth]{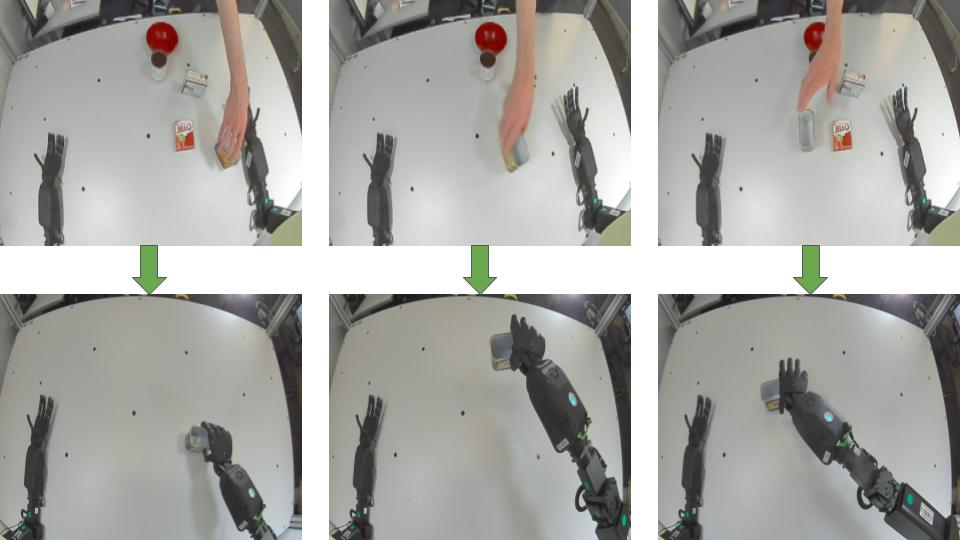}
    \caption{The top three images are taken from one of the human demonstrations at three timesteps: when the object is first grasped, during the movement, and after it is released. In the bottom three images, we can see three timesteps taken from the imitation of said demonstration. Again we show the moment when the object is first grasped, being moved, and after it is released. While the demonstration was given by a human sitting across from the robot, the model is able to perform the action from the robot's perspective. }
    \label{fig:enter-label}
\end{figure}
However, while having the potential to solve many problems, imitation learning also comes with difficulties. To imitate someone, the ability to perceive and understand the actions done by the target of the imitation is necessary. One of the difficulties stems from the differences between the bodies of the imitator and demonstrator. One way to address this problem is to control the robot and use it as its own demonstrator \cite{mao2023learning,copete2016motor}. Another approach to simplify the action perception is to do the action from the perspective of the robot; since human data is easier to collect than data from a controlled robot \cite{wang2023mimicplay}. \\
Imitation learning can also be accomplished by making the goal of the action simpler to understand. This way, the variance of actions is reduced, making the imitation process easier. This could be done by using language to specify the tasks in one demonstration, limiting the possible actions from the start on, or using pre-learned knowledge about the objects that will be part of the imitation \cite{shridhar2023perceiver,lopes2007affordance}.\\
Simplifying the imitation often means we lose out on some capabilities such as generalisability or that we need more pre-existing conditions to be fulfilled in order to allow the approach to work.
We do not try to simplify the imitation but the action instead. For this, we use demonstrations from a human from a different perspective, so a demonstration that could easily happen in real life and the kind of imitation that we see naturally. This means that following the exact motion of the actions is close to impossible because of different bodies and different spatial positions in regard to potential objects that have to be moved. We deal with these challenges in two steps. First, we use action segmentation to find out step by step which actions have been executed by the demonstrator to achieve their goals. Our action segmentation model is based on a generative diffusion model. With the demonstration as our condition, the model generates action labels for each frame. We focus on the base actions that were found in the mirror neuron: grasping, holding and manipulating an object \cite{gallese1998mirror}. The results from the action segmentation are the action sequences the robot needs to replicate to imitate the demonstration. The next step is to use the knowledge of when the actions happen and with which objects they happen to find out where we need to perform the actions. As we know the time points in which the demonstrator performs their grasps, we can use the object detection on key frames to find out the 3D coordinates of where the objects that were grasped are. Combining the spatial and temporal knowledge of the actions, we can replicate them with the robot. So, from a simple demonstration of a complex action, we extract the knowledge of the order of actions as well as the goals of the actions and allow the robot to replicate them. To facilitate the movement, we use inverse kinematics which represents the body schema of our robot.\\
Our approach uses an uncommon strategy to facilitate imitation learning directly from human demonstration by simplifying the action instead of the imitation.


\section{Related Work}
\subsection{Imitation Learning}
Imitation learning is used in many forms and applications because of a few key advantages. While many machine learning approaches rely on the availability of large amounts of training data, imitation learning reduces this requirement by a large margin. By using the demonstrations from an expert who is already capable of the task at hand, far less training data is needed. Expert demonstrations also simplify how to represent an action. For many tasks we deal with in our daily lives, it is much easier to show how to do them rather than explain how to do them. If we were to explain how we walk and how we move each of the muscles in our legs, the explanation would be far more contorted and inaccurate compared to simply showing how we move our legs. This aspect of imitation learning is especially helpful when it comes to robotics, where many tasks of such a nature are encountered. The high dimensionality that many motion tasks for humanoid robots require is far easier to express through the use of demonstrations. Therefore, robotics and imitation learning often go hand in hand \cite{schaal1999imitation}. Of course, getting an expert to do demonstrations to learn from is a challenge in itself which is why first attempts tried to learn quickly from single demonstrations \cite{atkeson1997learning}. This failed at the time, in part due to the differences between humans and robots, as well as the mechanisms controlling the robot being imprecise. The task chosen, swinging a pendulum, might also have been particularly difficult because of the rapidly shifting forces that are only indirectly controlled. However, kuniyoshi et al. \cite{kuniyoshi1994learning} showed that learning from a single demonstration is possible. They had to focus on one task and forgo a lot of generalisation, but through using a strategy similar to our approach of finding action sequences and using the perceived body-object relations, they managed to show that learning by watching and teaching by showing can work. 
Another way to get demonstrations is to hard code the actions for the robot and then use the recorded execution as a demonstration to learn from through imitation learning \cite{mao2023learning}. This facilitates the creation of many demonstrations. Instead of depending on the human demonstrator, it is, however, dependent on the flexibility of the written script in order to get a range of demonstrations. To add to the scripted generation, language commands can be used to further direct the task\cite{elshaw2004associator, shridhar2023perceiver}. Using a scripted robot for the demonstrations can make it easier to record more modalities than just vision \cite{hwang2017predictive}. Using a simulation makes it possible to reduce the demonstrations to more important parts, such as only the end-effector, which can improve the imitation learning \cite{lu2023decoupling}. Mixing human and robotic demonstration can give the benefits of both \cite{wang2023mimicplay} in turn needing both sorts of demonstrations and the capabilities to handle them. 

\subsection{Vision}
\subsubsection{Action Segmentation}
To imitate a demonstrator, the robot needs to be able to perceive them. Without an additional sensor that can be attached to the demonstrator or using the robot as a demonstrator for itself, this perception relies on the visual sense of the robot. From the demonstration, the robot should perceive the actions that are done and where they happen. To identify the actions and their temporal connections, we use an action segmentation model. The temporal abilities of RNNs \cite{TBW16} or transformers have improved the action segmentation quite a bit, ASFormer (Action Segmentation transFormer) \cite{yi2021asformer} shows how to modify transformers to fit the task of action segmentation. They have improved how to handle the input sequence and how to better utilise the temporal information of the model. The ASFormer model can also easily be used as a backbone which has led to many modifications and uses of it. Another recently developed machine learning model is the diffusion model. In this generative model, a noise is put over the input and step-wise removed during inference. Apart from generating images, it has also been used for object detection \cite{chen2023diffusiondet}. From there, it has been further adjusted to be used in action segmentation. Using the ASFormer model as its backbone and features from the videos extracted with I3D \cite{carreira2017quo} as the condition, the model uses a noisy version of the ground truth as its input only to predict the ground truth as its output, removing the noise \cite{liu2023diffusion}. The diffusion process of having multiple differently noised versions of the input and the possibility of using multiple inference steps allows the model to improve the results of the ASFormer. Action segmentation can also be improved by using further modalities for the input. Instead of just RGB images, prompts can be used to give the network a direction. In addition to that, changing how the input looks and forming graphs out of the videos, therefore representing the data differently, can help improve the action segmentation \cite{zhang2022semantic2graph}. Having direct knowledge about the movement that happens during actions through the use of accelerometers is also helpful as the additional knowledge if fused correctly, leads to very good results \cite{van2023aspnet}. However, using a more complex representation of the data or more modalities also means that more work is necessary to get the action segmentation running and not all modalities will always be available, and therefore, limits the use of such methods. 

\subsubsection{Object Detection}
Apart from the temporal information, it is also important to gain a spatial understanding of the demonstrated actions. As the demonstrations are recorded in two-dimensional data, the first step of extracting spatial information is to find the two-dimensional position. As the robot acts in three dimensions, the third dimension is deducted in a second processing step. A single 2D image does not always offer enough spatial information to find the 3D coordinates of an object. Most 3D object detectors use three-dimensional data such as point clouds, voxels or depth cameras. While we only have one 2D image, the challenge in transforming the information to 3D is lower. Fortunately, all the objects are on the table, which mostly eliminates one of the dimensions. Traditional object detectors are trained on a specific set of class labels, stopping them from recognising objects outside of that set \cite{he2016deep}. However, first attempts with object detectors that are not dependent on a set number of classes have been made \cite{gu2021open}. 

\section{Imitating Human Action}
\begin{figure} 
    \centering
    \includegraphics[width=0.5\textwidth]{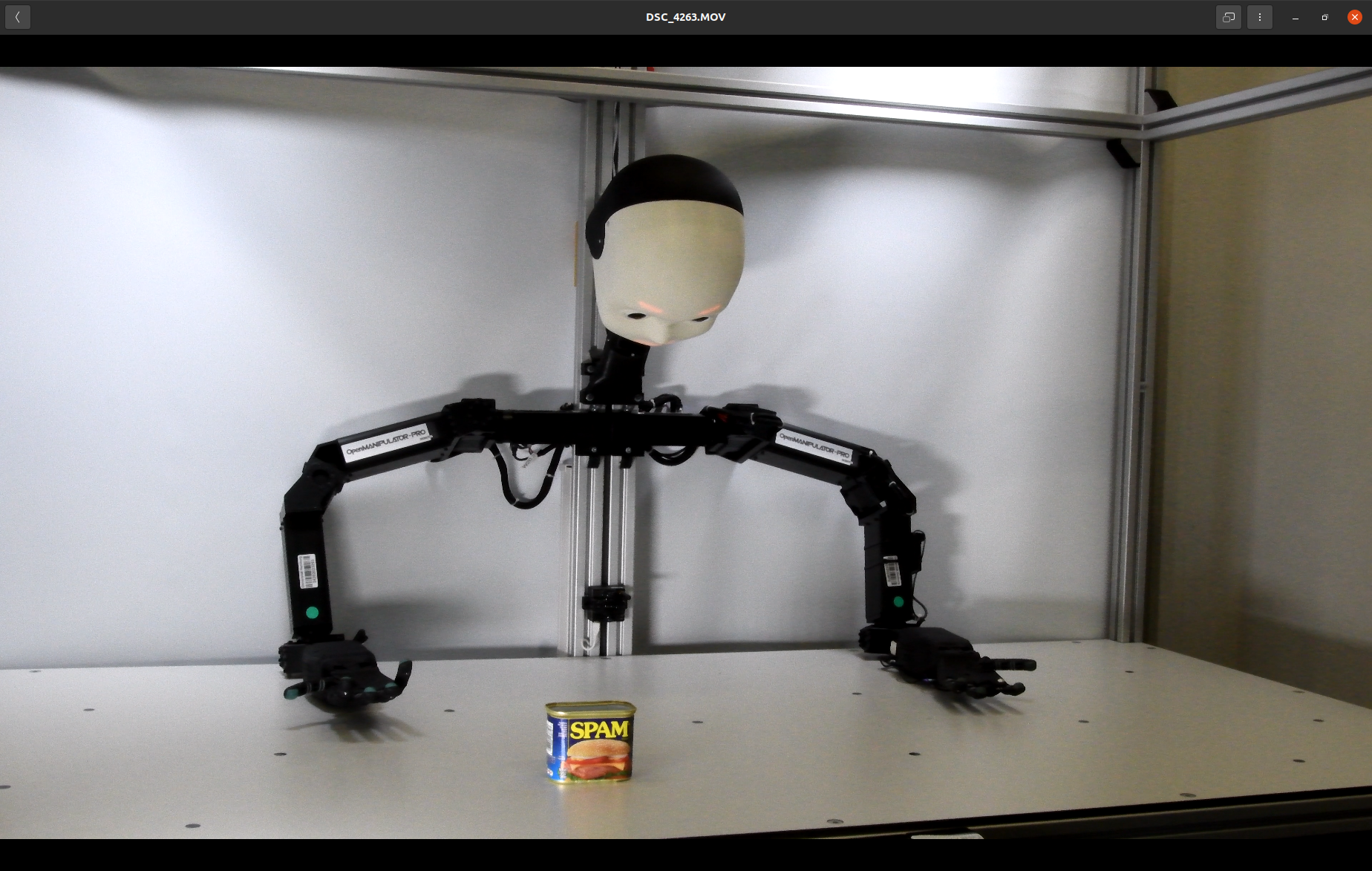}
    \caption{Our robot in its natural environment during one of the imitations where it grasps the spam can.}
    \label{fig:scene}
\end{figure}

\begin{figure*}[t] 
    \includegraphics[width=1\linewidth]{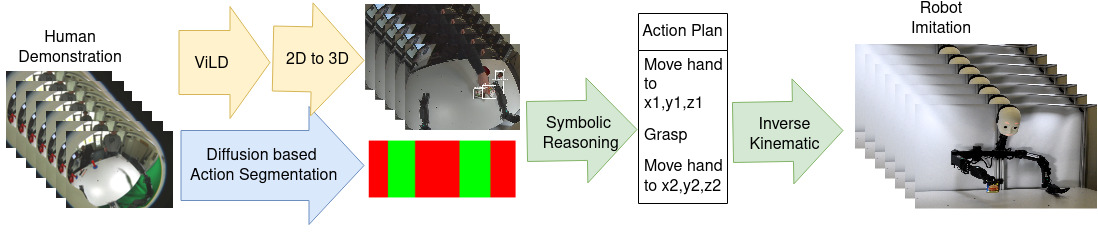}
    \caption{An overview of our architecture. The start of the approach is on the left, where the input is the human demonstration. We record the human demonstration as a video with the cameras in our robot's head. This video then goes through two models: our diffusion action segmentation model and the ViLD object detector. The object detector provides us with 3D positions for each detected object in each frame. The action segmentation model provides us with a class label for each frame. The class is either moving with the object or without. This information is used in our logical programming algorithm to create an action plan. This action plan is given to the inverse kinematic of our robot which then moves our robot to imitate the demonstration.}
    \label{fig:architecture}
\end{figure*}

The focus point of this approach is imitating human actions. We look at actions as hierarchical, with more complex actions, such as eating, being made up of a sequence of simpler actions, such as grabbing a fork, moving the fork to the food, picking up food and so on. Therefore, in our demonstration, we decided to demonstrate sequences of such simple actions that could be used in many situations. We pick up objects and move these objects before putting them down somewhere else. This sequence of grasping an object, moving it somewhere and releasing it is something that can be seen almost everywhere. For our scenario, we picked the workplace in front of our {NICOL} robot, as shown in Figure \ref{fig:scene}, so that the demonstrations happen in the same place as the imitations later on.
We use five objects, which are roughly hand-sized from the YCB object data set \cite{calli2015ycb}, the red bowl, the spam can, the Jello strawberry, the Jello chocolate cardboard boxes and the tomato can. For each of the objects, we record 24 videos in which we go through a sequence of picking one object up, moving it, releasing it, moving our hand away before picking it up once more, moving it to yet another place and releasing it again. We pick the objects up naturally, so the hand configuration differs between many attempts and the object is often occluded during grasping and being moved. Every scene holds multiple objects to simulate a cluttered working environment. The videos all have the same length of 20 seconds, and the demonstrations are all done by the same actor. \\
The part of our approach to detect and segment the actions from the demonstrations is based on the diffusion action segmentation model \cite{liu2023diffusion}. From the RGB video data, we create image flow images and use them as well as the RGB data to extract feature vectors using the I3D model \cite{carreira2017quo}. This is the part of our input that is used as the so-called conditioning in our diffusion model, similar to what is proposed by Ho et al. \cite{ho2020denoising}. The noised ground truth makes up the rest of the input. During training, the amount of noise is chosen randomly and applied to the ground truth and the amount of noise used is embedded in the input. This enables the network to iteratively denoise the output during inference. In diffusion models, there generally is a choice between predicting the noise and the ground truth. Our model predicts the ground truth, so we can skip some of the iterative denoising steps. Predicting the ground truth also allows us to use the cross entropy loss to directly see how good the predictions are. We tried all the losses used by Liu et al. \cite{liu2023diffusion} but found the best results when we only used a singular cross-entropy loss. For us, the most important aspect is correctly identifying the sequence of actions, while the exact time of transitions between actions is not quite as important. As the boundary alignment loss and the temporal smoothness loss direct the network to improve the correct classification of the borders of the actions, they are not beneficial here. Our backbone is based on the ASFormer model \cite{yi2021asformer}. The model has three encoder blocks and three decoder blocks. All of the blocks use sliding attention, and there are 28 layers in total. The output of our model has one class vector for each frame of the input video.\\
We also perform object detection on every frame of the video demonstration. For this, we are using ViLD \cite{gu2021open}. ViLD utilises a number of possible prompts representing possible parts of our scene and a pre-trained ResNet-50 which takes care of the region proposals. ViLD finds the text embeddings closest to the visual embedding and thus decides on a class. Our approach uses both of these information streams, combining them with symbolic reasoning in order to plan which action we need to do at what time and where to imitate the human demonstration. The first part of this is to use the knowledge gained about the action sequences to find the time steps in which the position of the objects is important for our actions. Then, the algorithm uses the object detection to find the positions of the objects in the key moments. We translate from two-dimensional coordinates to three-dimensional coordinates using our knowledge of the robot's camera position and pre-existing knowledge about the height of the table on which the demonstrations are performed. As the time steps often span over multiple frames, we use majority voting to improve the object detection. The majority voting starts from the frame in the key moment, checks how plausible the detected objects are, and, if necessary, compares it to the adjacent frames to increase the certainty of the detections. Using the knowledge of the positions of each object during the different actions, the algorithm determines which object is important for the actions and finds the key positions for each of the actions. Determining where to grasp the object and where to release it. Our approach utilises this knowledge to create an action plan. Finally, an inverse kinematic algorithm realises these abstract plans. \\
By concentrating on very basic human actions such as grasping, moving and releasing objects, we allow our approach to stay flexible and less dependent on specific actions. This is further emphasised by using an object detector that allows for many different object classes. The object detection came with two challenges: many false positives, and while the detections were mostly correct, the classifications were not quite as accurate. As the object detection was not fine-tuned, it often detected the robot or human arm as a further object. Neither of these detections was constant, however, making it harder to tell which object would be moved in a given action. This was further emphasised by our second challenge, which came from the classifications not always being consistent. This meant that in one frame, an object could be classified as a cardboard box and in the next one as a can. We use an off-the-shelf object detector to keep our approach generalisable and easy to set up with a pre-trained object detector. As the actions all went on for multiple frames, we used majority voting to narrow these problems down. When our detection was not concrete about which object was part of the action, the algorithm would continue to go to the neighbouring frames up to a threshold of 10 frames or until it got to a concrete decision. This helped with both of our challenges as it eliminated many of the false positives and allowed us to keep track of the objects even through miss-classifications. Further adjustments to the text prompts could improve the classifications.\\
The action segmentation only has to identify whether an action is actually happening or not, as an object in our scenario can only be manipulated after we grasp it and only be released after we manipulate it. The biggest challenges for the action segmentation came from the data set. As it consisted of natural human actions for the demonstrations, some of the grasps would completely obscure an object. Coupled with the distractor object, which also could occlude the object of the actions, some of the demonstrations were hard to follow for our model. The second part stems from most of the demonstrations following some natural patterns in regard to how long it would take to grasp an object or move it. In some of the demonstrations, however, the demonstrations escaped these patterns. This could, for example, lead to unusually long grasping actions or long times of inaction. These troubles could likely be reduced by increasing our data set. \\
As our action planning makes use of both of these sometimes flawed information streams, it is not always able to plan the action accurately. We can realise that our plan is flawed in many cases. There are demonstrations where multiple actions were detected correctly, but the object detections were flawed. This means the planner knows that there should be an action happening but does not know where the action happens or has conflicting information about where the action happens. There are also some cases where the object detection can let us know that the action segmentation is likely to be wrong, such as the action segmentation detecting an action that moves an object, but the object detection detecting the same positions as before the action. This knowledge could be used to inform the human working with the robot that it needs a repetition of the demonstration. A new demonstration should allow us to gain more correct information about the action and then realise that action.\\
As this approach always uses the same grasping technique, regardless of the object, it does fail for some object shapes. Of course, some other issues can also appear due to the rigid grasping technique, such as failing to grasp the object if it fell to one of its sides after it was released. We focus on directly imitating the demonstrator's goals and, therefore, learning a task extremely quickly. We are only using one demonstration for each sequence of actions, and directly using the body schema of the robot for grasping is one of the key factors. 
\section{Results}
\begin{figure}
    \centering
    \includegraphics[width=\linewidth]{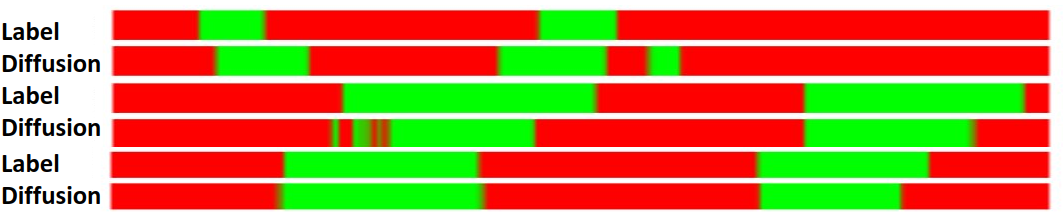}
    \caption{The results from the action segmentation on three demonstrations. For each demonstration there is one bar showing the ground truth and one bar showing the results from the action segmentation. For each of them, the time goes from left to right. The two possible classes are shown with different colours: red is for movement without an object, and green is for movement with an object. }
    \label{fig:diffusionResults}
\end{figure}

The nature of our approach allows us to see in detail to which degree the individual parts perform. First, we start with the action segmentation, where we look at both the direct prediction as well as after 100 diffusion steps. As we directly predict the ground truth instead of the noise, we do not have to go through all the time steps during inference. We randomly partition our data into a train set and a test set with 80\% of the samples for training and the other 20\% for testing. In Table \ref{tab:resultTable}, we can see the results from using different combinations of losses. We reach an accuracy of almost 90\% for the action segmentation. This accuracy is classification per frame, and for each video sample, we had 440 frames. The accuracy seems quite high compared to some of the benchmark data sets, on which the best performances tend to be in the mid-eighties. However, we only have three classes, so it can be expected that our model does have a slightly higher accuracy on our data set. Apart from the overall accuracy, we can also see that the model seems to have a few videos on which it performs quite badly, approaching a 50\% accuracy. These are often the video samples in which the object is very hard to see, as it has low contrast compared to the background, or in which the hand grasping the object significantly occludes the object. We show the results of the action segmentation model for three demonstrations in Figure \ref{fig:diffusionResults}. In the first example, the model mistakenly detects a third grasp and movement of the object. The second example has a very good performance, with only the borders of the segmented actions not quite matching up, while in the third example, we see this effect in higher intensity with some uncertainty in the first grasping action the diffusion detected. The red parts of the bar are when the label should be that the demonstrator's hand is moving on its own, while the green label represents the action starting with grasping an object until it is released. During the annotation, we counted the action as starting with grasping when all fingers involved in the grasp touched the object, and similarly, once this was not the case anymore, it would count as released.
\begin{table}
    \centering
    \begin{tabular}{c|c|c|}
         Losses & Action Segmentation & Position Detection \\
         CE & 0.8919  & 0.8171 \\
         CE +BA & 0.8732 &0.7777 \\
         CE + BA + TS& 0.8748 &0.7453 \\
         CE + TS & 0.8778 &0.7314
    \end{tabular}
    \caption{This table shows the results of our approach. The first column denotes the losses used for the action segmentation. CE = cross-entropy, BA = Boundary Alignment, TS = Temporal Smoothness. The second column shows the accuracy of the action segmentation, and the third shows the accuracy in determining the 3D positions.}
    \label{tab:resultTable}
\end{table}
Imitating Human Actions

The final positions obtained from the object detection have an accuracy between 0.73 and 0.81, as shown in Table \ref{tab:resultTable}. We have used a pre-trained object detector with an open vocabulary, but the classification has mostly been suboptimal. The classification is based on predefined possible classes; we used 15 possible classes, including "metal can" for the cans, "cardboard box" for the Jello boxes, and "red plate" for the red bowl, as "bowl" did not seem to work well and many more. The network ended up classifying a large majority of all the objects as cardboard boxes. The object detection part went better than the classification. The main problem that we encountered here was the irregular detection of the robot arm. While most objects were consistently detected, the robot arm was only detected in some of the frames. The second challenge that the object detection had was that objects would be obscured during the grasping action, making it harder to detect them. Our object detector also tracks objects over time which helps in detecting the movement of said objects. This mechanism is also hindered by the aforementioned challenges. Despite these challenges, we are able to use the RGB data as well as the camera position and the known table position to accurately find the 3D coordinates, where the actions take place.
To test the object detection directly, we randomly took 500 images of our data set; we included some knowledge for the object detection, mainly that we have five objects we want to detect. The objects are detected with an accuracy of 0.8724 and classified with an accuracy of 0.5011. The object detector also detects parts of the image as objects wrongly, with an average of 0.406 wrong objects detected per image.\\
When it comes to using the approach in the real world with our {NICOL} robot, a few more difficulties present themselves. Some of the objects we use are quite hard to grasp; both the red bowl and the tomato soup are round and need a specific approach and angle to grasp them. For this reason, we concentrated on the spam can and both of the Jello boxes. The next challenge is the positioning of the objects. We are using the images from the demonstration and rebuild the scene. As we have also only focused on the object that is the target of our action, the approach does not consider possible obstacles that we have to avoid. To make sure that no accidents happen, we rearrange the distractor objects to make sure they are not in the trajectory of the robot arm. Then, we use the demonstration fitting the current setup to generate our action plan and execute said plan. \\
Our grasps have a success rate of 69,44\%. As we had two grasps per demonstration, we put the object into the correct position if the first grasp failed. Most failures with the grasps came from inaccuracies in the exact position. This means that the robot's finger did not correctly connect to the object, causing it to be pushed to the side. The first grasp had a success rate of 66\% while the second grasp had a success rate of 72\%. We only used demonstrations where the generated plan was correct to avoid accidents. In these cases, the imitation worked out completely with a 44\% success rate.

\section{Conclusion}
In this paper, we show that robots can learn how to perform tasks directly from human demonstrations despite different bodies and perspectives. As these demonstrations are very natural and easy to collect and we only need a small number of demonstrations, this allows the robot to quickly adapt to a new task. Our approach demonstrates the benefits of learning by observing: what would otherwise be a hard-to-describe solution to a task can be demonstrated by actions instead. Despite using a small number of demonstrations and a large difference between the demonstrations and desired imitations, we achieve a high success rate in performing pick-and-place tasks. Our approach only takes in the demonstrations and only uses these demonstrations to learn. The ability learned only from one demonstration is enough to successfully imitate the demonstrator and achieve the demonstrated goals. This approach shows how powerful demonstrations are as a tool for learning.

\bibliographystyle{ieeetr}
\bibliography{literature.bib}

\end{document}